# Natural-Logarithm-Rectified Activation Function in Convolutional Neural Networks

YANG LIU, JIANPENG ZHANG, CHAO GAO, JINGHUA QU, AND LIXIN JI

National Digital Switching System Engineering and Technological R&D Center, Zhengzhou 450002, China

Corresponding author: Jianpeng Zhang (zjp@ndsc.com.cn)

This work was supported by the National Natural Science Foundation of China for Innovative Research Groups under Grant 61521003, and the National Natural Science Foundation of China under Grant 61601513 and Grant 61803384.

**ABSTRACT** Activation functions play a key role in providing remarkable performance in deep neural networks, and the rectified linear unit (ReLU) is one of the most widely used activation functions. Various new activation functions and improvements on ReLU have been proposed, but each carry performance drawbacks. In this paper, we propose an improved activation function, which we name the natural-logarithm-rectified linear unit (NLReLU). This activation function uses the parametric natural logarithmic transform to improve ReLU and is simply defined as $f(x) = \ln(\beta \cdot \max(0,x) + 1.0)$. NLReLU not only retains the sparse activation characteristic of ReLU, but it also alleviates the "dying ReLU" and vanishing gradient problems to some extent. It also reduces the bias shift effect and heteroscedasticity of neuron data distributions among network layers in order to accelerate the learning process. The proposed method was verified across ten convolutional neural networks with different depths for two essential datasets. Experiments illustrate that convolutional neural networks with NLReLU exhibit higher accuracy than those with ReLU, and that NLReLU is comparable to other well-known activation functions. NLReLU provides 0.16% and 2.04% higher classification accuracy on average compared to ReLU when used in shallow convolutional neural networks with the MNIST and CIFAR-10 datasets, respectively. The average accuracy of deep convolutional neural networks with NLReLU is 1.35% higher on average with the CIFAR-10 dataset.

## I. INTRODUCTION

Activation functions have a crucial impact on the performance and capabilities of neural networks, and they are one essential research topic in deep learning. Activation functions are generally monotonic and nonlinear. The introduction of the activation function in neural networks allows neural networks to apply nonlinear transforms to input data such that many complex problems can be resolved. However, as the number of neural network layers increases, serious problems caused by the activation function begin to appear. Some examples include gradient vanishing or explosion during back propagation and migration of the data distribution among layers of the neural network after activation, which increases the difficulty of learning the training data. Therefore, a suitable activation function is critical to building a neural network.

Rectified linear unit (ReLU) [1], [2] is one of the most widely-used activation functions in neural networks in recent years. In the majority of popular convolutional neural networks, e.g., VGG nets [3], residual networks (ResNets) [4], [5], and dense convolutional networks (DenseNets) [6], ReLU always provides desirable results. Meanwhile, the research community is still devoted to developing new activation functions (e.g., Gaussian error linear unit (GELU) [7], scaled exponential linear unit (SELU) [8], and Swish [9]) and improving ReLU (e.g., leaky ReLU (LReLU) [10], parametric ReLU (PReLU) [11], exponential linear unit (ELU) [12], and concatenated ReLU (CReLU) [13]) to obtain more robust activation functions. However, most new activation functions do not perform well in terms of generalized performance. Xu et al. [14] and Ramachandran et al. [9] systematically investigated the performance of different types of rectified activation functions in convolutional neural networks, and they presented inconsistent performance improvements with these activation functions across different models and datasets.

ReLU is still one of the most favorable activation functions in convolutional neural networks due to its simplicity, effectiveness, and generalization. However, the advantage of simplicity can also cause problems. ReLU cannot provide effective control over neuron activation and gradients within a reasonable range when the neuron's input $x > 0$. Also, ReLU has the characteristic of sparse activation. These characteristics constitute its major drawback, i.e., the "dying ReLU", as well as bias shift problems. Most activation functions for improving ReLU have abandoned the sparse activation characteristic to avoid the "dying ReLU" problem [8], [10]-[13]. However, sparse activation of the activation function plays a vital role in neural networks in terms of information disentangling [2]. Thus, in this paper, we propose a parametric natural logarithmic transformation to improve the $x > 0$ part of ReLU while retaining the sparse activation characteristic of the activation function, yielding an improved activation function named the natural-logarithm-rectified linear unit (NLReLU). We show that NLReLU can reduce heteroscedasticity in data distribution among layers and the bias shift effect, thus alleviating the "dying ReLU" and vanishing gradient problems to some extent, as well as improving the performance of the neural network.

The primary contributions of our work are summarized as follows:
- We propose an improved activation function based on the parametric natural logarithm transformation, namely NLReLU.
- We analyze the improvements provided by NLReLU compared with ReLU and investigate other characteristics and advantages of NLReLU.
- We conduct several experiments with convolutional neural networks of different depths and datasets, and the experiments show that our activation function provides performance that is better than ReLU and is comparable to other well-known activation functions.

This paper is organized as follows. In Section II we review work on related activation functions. Our proposal is described in Section III. Experimental results are reported and discussed in Section IV. Finally, conclusions are given in Section V.

## II. RELATED RESEARCH
In this section, we briefly review and analyze the advantages and disadvantages of some activation functions, which will be compared with NLReLU, as well as the evolution of these activation functions.

### A. SIGMOID AND TANH
The Sigmoid and Tanh functions are *S*-shaped activation functions that are continuously differentiable and smooth. Both were commonly-used activation functions in early neural networks [15]. The first derivative of the Sigmoid is large when the input $x \sim 0$, thus the output values from neurons are essentially pushed towards the extremes during the gradient update process. However, the gradient of the Sigmoid function is small overall, especially becomes smaller farther from $x = 0$, and the gradient of the Sigmoid function is prone to saturate near 0; this is the essence of the vanishing gradient problem. The definition of saturation of the activation function was previously described [17]. Tanh was proposed to solve the non-centrosymmetric problem encountered with the Sigmoid function [16]. However, the vanishing gradient problem still exists, because the gradients are also prone to saturate near zero. Saturation of the activation function hinders training of neural networks [17].

### B. RELU AND SOFTPLUS
Since the first time it was shown that deep, purely supervised networks could be trained [2], the robust performance of ReLU [1], [2] has become the default activation function for most convolutional neural networks. ReLU is a piecewise linear function, unlike the continuously differentiable S-shaped activation functions. ReLU is left-hard saturated, i.e., the output and the gradient are both 0 when the input $x < 0$. However, since the derivative is 1 when the input $x > 0$, the gradient of ReLU does not attenuate when the input is positive, thus effectively solving the vanishing gradient problem. ReLU has the characteristic of sparse activation. Glorot et al. [2] analyzed this sparsity and fund that it can provide neural networks with many advantages, e.g., information disentangling, efficient variable-size representation and linear separability. Unfortunately, ReLU-activated neurons are prone to death during training when the learning rate or the gradient is too high. Since the first derivative of ReLU is 1 and ReLU does not restrain the gradient, large gradients tend to cause neurons to fall into the hard saturation regime and "die."

Another problem with ReLU is bias shift [12]. When $x > 0$, ReLU lacks effective restraints on output values and gradients, which tends to cause the bias shift problem and large heteroscedasticity in neuron data distributions among layers in the neural network. This makes model training more difficult.

Softplus [1], [2] can be treated as a smooth fitting of ReLU. Softplus is a monotonic and strictly positive activation function. The first derivative of Softplus is a Sigmoid, and therefore Softplus is left soft saturated. Compared to ReLU, Softplus has a wider acceptance domain and is continuously differentiable, but the calculation of Softplus in neural networks is complicated. Meanwhile, ReLU performs better than Softplus in most cases, so smooth fitting does not improve ReLU, and ReLU is used more commonly.

### C. LRELU AND PRELU
LReLU [10] solves the problem of "dying ReLU" by giving negative values a non-zero slope. This slope is typically a small constant, e.g., 0.01. LReLU both corrects each hidden layer's data distribution and retains the negative values so that information associated with the negative values is not lost. However, LReLU's parameter must be manually defined based on prior knowledge, limiting the improvement provided by LReLU over ReLU [14].

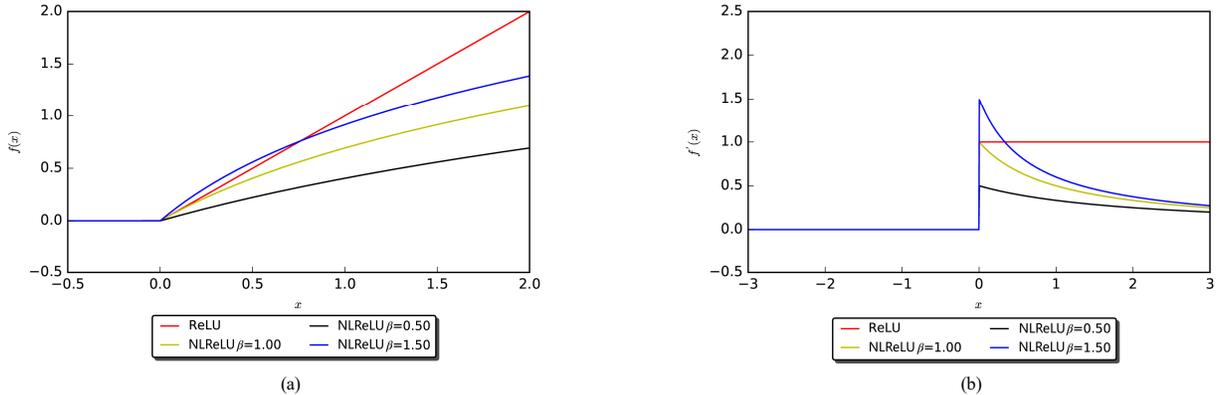

**FIGURE 1.** NLReLU and the first derivative of NLReLU: (a) NLReLU compared with ReLU, and (b) comparison of the first derivatives of NLReLU and ReLU.

PReLU [11] provides further improvements over LReLU. Compared to LReLU, the slope of the negative numbers in PReLU can be learned and are trainable rather than fixed. When the slope is very small, PReLU degenerates into ReLU or LReLU. PReLU also converges faster. Since the output from PReLU is closer to the zero-mean value, the stochastic gradient descent process is closer to the natural gradient [11].

LReLU and PReLU show that the performance of the activation function can also be effectively improved without sparse activation. However, their disadvantage is that they cannot ensure a noise-robust deactivation state [12].

### D. ELU, SELU AND SWISH

In contrast to LReLU and PReLU, ELU [12] and SELU [8] are both left-soft saturated. The linear part of ELU and SELU allows them to avoid the vanishing gradient problem, while the left-soft saturation makes ELU and SELU more robust for input noise or variations. The advantage of ELU and SELU is that mean activations can be pushed closer to zero. Clecert et al. [12] showed that the mean shifts towards zero speed up the learning process by reducing bias shift.

Swish [9] is inspired using a Sigmoid function as the gating in long short-term memory neural networks. The Swish function can be treated as a smooth function classified as somewhere between ReLU and a linear function. Swish is unsaturated and non-monotonic, and some empirical results show that the use of Swish in deep convolution networks is better than the use of ReLU [9].

### III. PROPOSED METHOD

ReLU must face the existing problems of "dying ReLU" and bias shift. In Section III, we find that the aforementioned activation functions (LReLU, PReLU, ELU, and SELU) can improve upon the disadvantages of ReLU as they do not require sparse activation. Without sparse activation, the problem where neurons are prone to death can be avoided; keeping negative numbers and striving to shift the mean activations to 0 alleviates the bias shift effect.

However, the nature of sparse activation is important for the activation function, and rectifier activation functions allow networks to easily obtain sparse representations. A certain degree of sparsity contributes to the mathematical advantages of the network, e.g., information disentangling and linear separability. Thereby, we want to improve upon the problems faced by ReLU while retaining the characteristic of sparse activation. We introduce a parametric natural logarithmic transform to improve the $x > 0$ portion of ReLU. NLReLU is defined as follows:

$$f_{\text{NLReLU}}(x) = \ln(\beta \cdot \max(0, x) + 1.0) \quad (1)$$

The first derivative of NLReLU is:

$$f'_{\text{NLReLU}}(x) = \begin{cases} \dfrac{\beta}{\beta \cdot x + 1} & , \text{if } x \geq 0 \\ 0 & , \text{otherwise} \end{cases} \quad (2)$$

Fig. 1a shows a plot of NLReLU for different values of $\beta$. The first derivative of NLReLU is shown in Fig. 1b for different values of $\beta$. The concept of NLReLU was first used in attention-gated convolutional neural networks [18] for sentence classification, where a standard logarithmic transform was used to make the neural network more sensitive to small attention weights. In this paper, we introduce a natural logarithmic transformation with the parameter $\beta$ to improve ReLU. The scale of $\beta$ controls the speed at which the response of the activation function rises as the input grows. If $\beta < 1$, the function value and derivative of NLReLU are both less than that of ReLU. A larger $\beta$ value means that NLReLU neurons are more sensitive to small values because the activation response rises faster. The mean activation is closer to zero when $\beta$ is smaller. $\beta$ increases the adaptability of NLReLU and allows it to be fine-tuned to different neural networks.

A logarithmic transformation is often used in econometrics to convert skewed data before further analysis [19]. Since the raw data from the real world is often heteroscedastic, remedial measures like a logarithmic transformation can decrease this heteroscedasticity. Thus, the data can be analyzed with conventional parametric analysis.

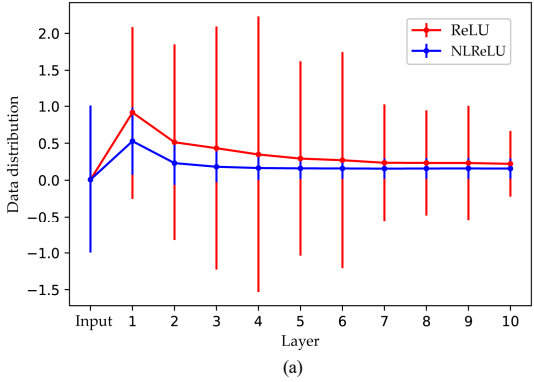 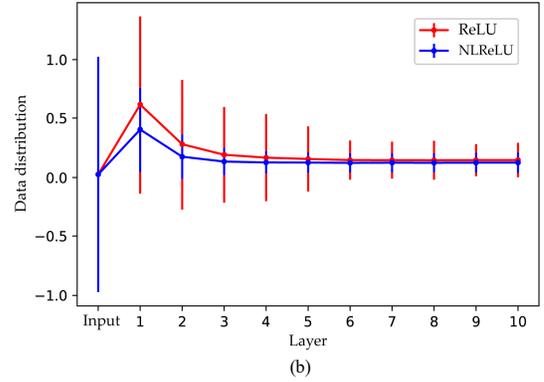

**FIGURE 2.** Simulation based on a fully-connected neural network with 10 hidden layers and 100 neurons per hidden layer. The batch size of the input data is 100. The input data is standard normally distributed, the bias term of each activation is initialized as 0.1, and the weights are initialized as: (a) normal distribution $N(0, 1.5)$, and (b) standard normal distribution $N(0, 1)$. We simulated the case where the network has large heteroscedasticity among layers. We recorded the mean and standard variance of the number of activated neurons of each layer.

Each hidden layer in the ReLU network is easy to bias shift, and the distribution of activated neurons between each layer of the network is prone to be heteroscedastic. Although these problems can be improved by improving the initialization method (e.g., Xavier [20], MSRA [11]), ReLU still cannot effectively control the change in the data distribution during training. Meanwhile, Li et al. [21] report that it is difficult for the 30-layer ReLU network to converge if batch normalization (BN) [22] is not used; this arises even if the network is initialized with Xavier [20] or MSRA [11]. Introducing the logarithmic transformation into the activation function can shift the mean activation in each layer closer to 0 while reducing the heteroscedasticity in the data distribution among layers. As shown in Fig. 2, we simulate the case where the network has large heteroscedasticity among layers, and one can see that NLReLU converts the mean activations per layer to approximately "normal" and effectively reduces heteroscedasticity among the layers.

Clevert et al. [12] analyzed the relationship between the bias shift effect and the mean number of activations, and they proved that reducing the bias shift effect is equivalent to pushing the mean number of activations close to 0 such that the stochastic gradient descent process will be closer to the natural gradient descent, thereby reducing the number of iterations required and accelerating the learning speed (see Theorem 2 and text thereafter in [12]). As demonstrated in Fig. 2, NLReLU can speed up the learning process because NLReLU can shift the mean activations closer to 0 and reduce the variance more than ReLU; thus, reducing bias shift.

The calculation of backpropagation and gradient based in NLReLU is simple. We assume that $n$ is a neuron with parameter vector $w$ and bias term $b$ in a hidden layer of the neural network. $x$ is the input of $n$ and $y$ is the output. $L$ is the current loss of the network. Based on the chain rule, when using NLReLU, the gradients of the weight $w$ and bias $b$ are:

$$\begin{cases} \nabla_w L = \nabla_y L \cdot \nabla_w y = \nabla_y L \cdot \dfrac{\beta x}{\beta \cdot \max(0, wx+b)+1} \\ \qquad = \nabla_y L \cdot f'_{NLReLU}(wx+b) \cdot x \\ \nabla_b L = \nabla_y L \cdot \nabla_b y = \nabla_y L \cdot \dfrac{\beta}{\beta \cdot \max(0, wx+b)+1} \\ \qquad = \nabla_y L \cdot f'_{NLReLU}(wx+b) \end{cases} \quad (3)$$

However, the gradients of weight w and bias b when using ReLU are:

$$\begin{cases} \nabla_w L = \nabla_y L \cdot \nabla_w y = \nabla_y L \cdot x \\ \nabla_b L = \nabla_y L \cdot \nabla_b y = \nabla_y L \cdot 1 \end{cases} \quad (4)$$

From (3) and (4), the first derivative of NLReLU has an effect on the magnitude of the gradient compared with ReLU. According to (2) and Fig. 1b, the first derivative of NLReLU decreases as the input increases, and NLReLU is right-soft saturated. Empirical results show that $\beta$ generally ranges from 0.7 to 1.1 (see Section IV).

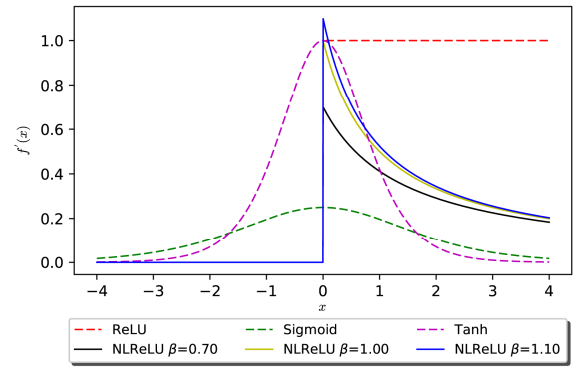

**FIGURE 3.** First derivatives of ReLU, Sigmoid, Tanh, and NLReLU for different values of $\beta$.

As plotted in Fig. 3, because the gradient is only slightly magnified by NLReLU's first derivative in a very small interval when $\beta > 1$, NLReLU could minify most gradients and reduce the risk of neurons dying due to large gradients. Meanwhile, as shown in Fig. 3, although NLReLU is right-soft saturated, the first derivative of NLReLU is less likely to fall into the saturation regime where the derivative is very close to 0 compared with Sigmoid and Tanh, thereby alleviating the vanishing gradient problem.

NLReLU could alleviate the "dying ReLU" and vanishing gradient problems to some extent, thereby improving convergence of neural networks. Experiments show that NLReLU networks can still converge when the learning rate is increased, whereas ReLU networks cannot converge (see Section IV-A).

While the use of NLReLU narrows neurons via the parametric logarithmic transformation, it improves discrimination of small activation values. Moreover, this phenomenon is more pronounced as $\beta$ increases. For example, the difference between 0 and 0.25, and between 0.75 and 1, are both 0.25 when they pass the NLReLU neuron with $\beta = 0.7$:

$$\begin{aligned} \ln(0.7 \times 0.5 + 1) - \ln(0.7 \times 0.25 + 1) &\approx 0.14 > \\ \ln(0.7 \times 1 + 1) - \ln(0.7 \times 0.75 + 1) &\approx 0.11 \end{aligned} \quad (5)$$

Meanwhile, when they pass the NLReLU neuron with $\beta = 1.0$, the relative distinction of small activation values is more obvious, i.e.:

$$\begin{aligned} \ln(1.0 \times 0.5 + 1) - \ln(1.0 \times 0.25 + 1) &\approx 0.18 > \\ \ln(1.0 \times 1 + 1) - \ln(1.0 \times 0.75 + 1) &\approx 0.13 \end{aligned} \quad (6)$$

That is to say, the parametric logarithmic transformation makes NLReLU more sensitive to differences in most small activation values, and effects of some larger values are weakened (e.g., those due to noise in the data).

The advantages of NLReLU can be summarized as follows:
- NLReLU can be fine-tuned to different networks.
- NLReLU can push the mean activation of each hidden layer close to 0 and reduce the variance, as well as reduce heteroscedasticity in the data distribution among layers and the bias shift effect, thereby speeding up the learning process.
- NLReLU minifies most gradients and helps prevent gradients from falling into the saturation regime, therefore it helps solve the "dying ReLU" and vanishing gradient problems to some extent, thus improving convergence.
- NLReLU provides more obvious discrimination of small activation values, thus weakening the effect of some large activation values (e.g., due to noise in the data).

## IV. EXPERIMENTS

We have undertaken intensive studies involving convolutional neural networks with different depths and architecture to compare the effectiveness and generalization of different activation functions. These studies were conducted with two essential datasets, i.e., MNIST[1] [23] and CIFAR-10[2] [24]. We categorize these networks into two groups in terms of their depths, i.e., shallow and deep convolutional neural networks. Shallow convolutional neural networks have less than 19 convolutional layers, while deep convolutional neural networks have 50 to 200 convolutional layers. The network implementations we used are provided by Tensorflow [25].

We emphasize that our aim here is not to constantly use various tricks (e.g., moderate data augmentation) or fine-tune parameters to improve on the state-of-the-art results. Rather, the primary purpose of this paper is to verify and evaluate the effectiveness of the proposed method in a situation without the influence of other tricks. We use the 10-fold cross-validation method on the training set to determine suitable parameters that provide effective network convergence. We then replace ReLU with other activation functions. Each experiment was run 10 times, and the mean and standard deviation were recorded as the final result. All the results in this paper are presented with error bars.

### A. SHALLOW CONVOLUTIONAL NEURAL NETWORKS

In this section, we conduct a series of comparative experiments with the SimpleCNN (a simple convolutional neural network with three weighted layers), LeNet-5 [23], AlexNet [16], and VGG networks [3], including VGG Net-A (VGG-11), VGG Net-D (VGG-16), and VGG Net-E (VGG-19). All comparative experiments were performed with the MNIST and CIRAR-10 datasets.

#### 1) EXPERIMENT SETUP

SimpleCNN is a simple convolutional neural network that we constructed with only one convolution layer. SimpleCNN has only one convolutional layer with 64 convolutional kernels of size $5 \times 5$, one fully-connected layer with 1024 nodes, followed by a softmax output layer. LeNet-5 and AlexNet were configured with the original settings [23], [16].

All network weights were initialized according to the procedure in Xavier [20]. Biases were initialized as 0, the Adam optimizer was used for all networks [26] and the learning rate $lr$ is $1 \times 10^{-4}$. The batch size with the MNIST dataset was set to 100, and we trained SimpleNN and LeNet-5 for 30,000 iterations, while AlexNet was trained for 10,000 iterations. Regarding the CIFAR-10 dataset, the batch size was set to 250 and 100 for AlexNet and VGG networks, respectively; AlexNet and the VGG networks were trained for 30 epochs (about 6,000 iterations) and 120 epochs (about 60,000 iterations), respectively. The keep-rate was 0.75 for AlexNet, and 0.5 for the VGG networks. The selection of parameter settings and training iterations is based on the fact that the model can smoothly converge in 10-fold cross-validation without overfitting. The other parameters were maintained at their original values [23], [16], [3].

---

[1] http://yann.lecun.com/exdb/mnist/
[2] http://www.cs.toronto.edu/~kriz/cifar.html

**TABLE 1.** Classification accuracy (%) of shallow convolutional neural networks with the MNIST test set using different activation functions.

| Activation Function | SimpleCNN | | LeNet-5 | | AlexNet | |
|---|---|---|---|---|---|---|
| | $lr = 10^{-2}$ | $lr = 10^{-4}$ | $lr = 10^{-2}$ | $lr = 10^{-4}$ | $lr = 10^{-3}$ | $lr = 10^{-4}$ |
| NLReLU | **98.61±0.23** ($\beta$=1.00) | **99.26±0.05** ($\beta$=1.00) | **98.96±0.09** ($\beta$=1.00) | **99.44±0.04** ($\beta$=0.95) | 99.38±0.03 ($\beta$=1.00) | **99.48±0.02** ($\beta$=1.05) |
| ReLU | —— | 99.13±0.03 | —— | 99.22±0.04 | —— | 99.34±0.01 |
| Softplus | —— | 98.43±0.09 | —— | 99.02±0.03 | —— | 99.10±0.07 |
| Swish | —— | 99.04±0.04 | —— | 99.24±0.03 | 99.28±0.03 | 99.32±0.04 |
| LReLU | 93.93±0.98 | 99.14±0.02 | 98.12±0.20 | 99.31±0.02 | **99.46±0.04** | 99.39±0.03 |
| PReLU | 97.83±0.22 | 99.17±0.03 | 98.11±0.14 | 99.34±0.03 | 99.43±0.09 | 99.38±0.04 |
| ELU | —— | 99.19±0.05 | —— | 99.38±0.01 | 99.42±0.02 | 99.44±0.01 |
| SELU | —— | 99.21±0.05 | —— | 99.36±0.07 | 99.44±0.03 | 99.44±0.02 |

**TABLE 2.** Classification accuracy (%) of deeper shallow convolutional neural networks with the CIFAR-10 test set using different activation functions.

| Activation Function | AlexNet | VGG-11 | VGG-16 | VGG-19 |
|---|---|---|---|---|
| NLReLU | **70.56±0.15** ($\beta = 1.00$) | **77.35±0.36** ($\beta = 0.70$) | 79.94±0.31 ($\beta = 0.85$) | 80.53±0.25 ($\beta = 0.85$) |
| ReLU | 68.04±0.28 | 75.32±0.44 | 78.29±0.21 | 78.56±0.23 |
| Softplus | 67.91±0.53 | 75.07±0.35 | 77.82±0.47 | 78.11±0.32 |
| Swish | 69.19±0.40 | 77.04±0.31 | 79.09±0.26 | 80.05±0.51 |
| LReLU | 68.55±0.22 | 75.45±0.58 | 79.42±0.34 | 80.41±0.21 |
| PReLU | 68.84±0.37 | 76.09±0.41 | 79.31±0.38 | 80.36±0.45 |
| ELU | 68.89±0.20 | 76.20±0.35 | 79.81±0.29 | 79.94±0.22 |
| SELU | 69.31±0.27 | 77.13±0.20 | **80.18±0.37** | **80.74±0.29** |

2) RESULTS AND DISCUSSION

Tables 1 and 2 show the classification accuracy results for different activation functions and datasets with these shallow convolutional neural networks. As shown in Table 1, NLReLU provides 0.16% higher accuracy than ReLU on average. The performance of Swish in shallow convolutional neural networks is not ideal. The accuracy provided by LReLU and PReLU with shallow networks is similar. Both ELU and SELU perform better than LReLU and PReLU due to the reduced bias shift and their self-normalization characteristic. Since MNIST is a classic dataset, the classification accuracy of each convolutional neural networks is generally close to 100%, therefore the improvement in classification accuracy is relatively small. However, the improvement provided by NLReLU is significant compared to ReLU.

In the experiment with the MNIST dataset, we conducted another set of control experiment to explore the convergence of the networks with different activation functions as the learning rate increases. As shown in Table 1, the networks using ReLU, Softplus, and other activation functions cannot converge effectively as the learning rate increases, but using NLReLU can help improve convergence. Since NLReLU can minify most gradients and helps prevent gradient saturation, it can help alleviate the "dying ReLU" and vanishing gradient problems, thus improving the convergence of the network. When the learning rate increases, the neurons in the ReLU networks are prone to death because the gradients are too large. These dead neurons will not be updated during subsequent training, and the network becomes too sparse and will not converge. LReLU and PReLU can also cause the networks to effectively converge due to their identity for positive values, and they keep negative values to avoid the dying neuron and vanishing gradient problems. Swish, ELU, and SELU cannot make SimpleCNN and LeNet-5 effectively converge when the learning rate increases.

We then conducted an experiment using several deeper shallow convolutional neural networks with the CIFAR-10 dataset. As shown in Table 2, NLReLU outperforms ReLU by an average of 2.04% and up to 2.52% maximum, and NLReLU outperforms all other activation functions except SELU. Within AlexNet, NLReLU performs 1.25% better than SELU, and using NLReLU in the VGG networks provides performance that is comparable to SELU. Tables 1 and 2 show that using NLReLU in the convolutional neural networks with 1 to 19 convolutional layers provides stable performance compared to other well-known activation functions.

We fine-tuned $\beta$ in the NLReLU for different networks as shown in Tables 1 and 2. To illustrate the sensitivity of the model to $\beta$, we used AlexNet as an example to perform sensitivity analysis on $\beta$ based on the MNIST and CIFAR-10 datasets. The experimental results are plotted in Fig. 4, where each result shows the mean value from 10 replicates and is shown with an error bar. The value of $\beta$ was varied from 0.60 to 1.20 in intervals of 0.05. One can see that the variance of model classification accuracy is larger as $\beta$ increases. The most appropriate $\beta$ value in the neural network with a dataset can be obtained by such sensitivity analysis. From Fig. 4, we know that for AlexNet, the most appropriate $\beta$ value for NLReLU is 1.05 and 1.00 with the MNIST and CIFAR-10 datasets, respectively. Meanwhile, Tables 1 and 2 show that $\beta$ in the VGG networks ranges from 0.70 to 0.85, and $\beta$ in networks such as AlexNet and LeNet-5 is approximately 1.00.

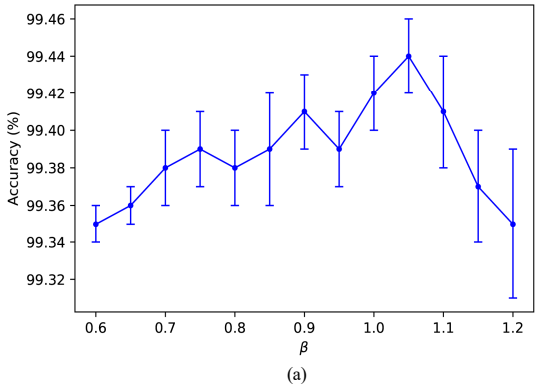 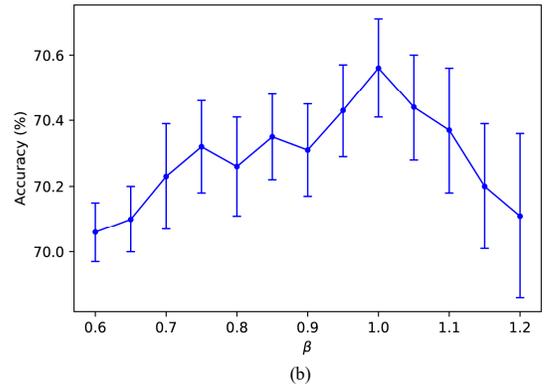

**FIGURE 4.** The impact of different $\beta$ on AlexNet's classification accuracy: (a) results of AlexNet experimented on MNIST, and (b) results of AlexNet experimented on CIFAR-10.

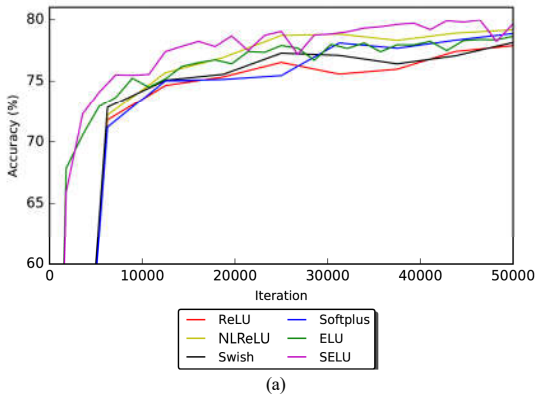 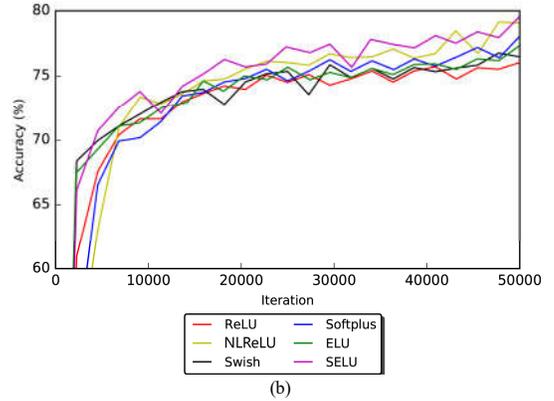

**FIGURE 5.** Training curves of VGG-19 on CIFAR-10. We randomly held out 10% of the training images to use as a validation set. (a) Training and (b) validation accuracy curves.

We also plot the learning curves for VGG-19 in Fig. 5. This figure shows that the accuracy with NLReLU initially increases slowly and later exceeds other activation functions after approximately 6,000 iterations. Furthermore, the accuracy provided by NLReLU is comparable to that provided by SELU.

### B. DEEP CONVOLUTIONAL NEURAL NETWORKS

In this section, we present results from a series of comparative experiments with the CIFAR-10 dataset using several deep convolutional neural networks, where the number of convolutional layers gradually deepened from 50 to 200, i.e., ResNet-50, ResNet-101, ResNet-152 and ResNet-200 [4], [5]. The ResNet we use here is the pre-activation variant proposed by He et al. [5].

#### 1) EXPERIMENT SETUP
The weights in each ResNet variant were initialized using the procedure defined in the MSRA [11]. The batch size was 128 and the Adam optimizer was used [26] with learning rate of 1 ×10⁻³. Other fine-tuning tricks, e.g., weight decay, learning rate decay, and data augmentation, were not used. The other parameter settings follow the original values [4], [5]. The parameters and number of training iterations was selected in order to ensure smooth convergence in 10-fold cross-validation without overfitting. We trained ResNets for 130 epochs (approximately 50,000 iterations).

#### 2) RESULTS AND DISCUSSION
An important and heavily-used structure in ResNet is the BN layer. The use of BN in the pre-activation variant of ResNet is accompanied by an activation function. The activation function is mostly used before the convolutional layer and after the BN layer, to "pre-activate" the input to the convolutional layer. Regarding the use of the activation function mainly placed behind BN in ResNet, we divide it into three categories according to its location, as shown in Fig. 6. The activation functions for position A are located after the BN layer of each residual unit's input. The activation functions for position B are the activation functions of each residual unit except position A. The activation functions for position C are located before the output of the entire ResNet.

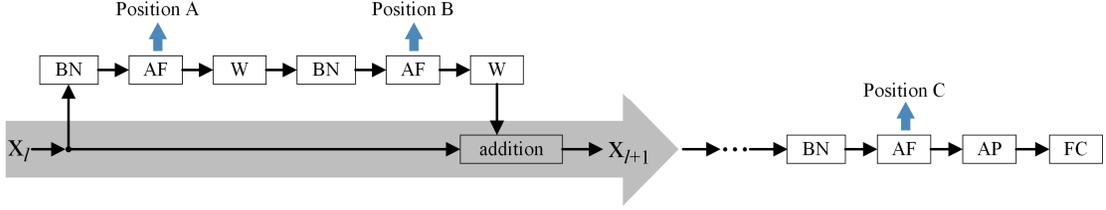

**FIGURE 6.** Example network architecture for the residual network, which is the full pre-activation residual network variant proposed by He et al. [5]. BN stands for batch normalization [22], AF stands for activation function, W stands for weight layer, AP stands for average pooling layer, and FC stands for fully-connected layer.

**TABLE 3.** Classification accuracy of ResNet-50 in different cases ($\beta = 1.00$) with CIFAR-10 dataset. We use a three-tuple consisting of 0 and 1 to represent different cases. The three-tuple (A, B, C) represents whether the three positions A, B, and C use the NLReLU, and a value of 1 indicates that the NLReLU of the corresponding position is kept. Other parameter settings follow Section IV-B-1.

| Condition | Accuracy (%) | Condition | Accuracy (%) |
| --- | --- | --- | --- |
| (0,0,0) | 73.55±0.78 | (1,1,0) | 76.95±0.55 |
| (1,0,0) | 75.77±0.47 | (1,0,1) | 81.32±0.42 |
| (0,1,0) | **82.34±0.26** | (0,1,1) | 81.69±0.37 |
| (0,0,1) | 80.23±0.35 | (1,1,1) | 80.74±0.45 |

**TABLE 4.** Classification accuracy (%) of deep convolutional neural networks with the CIFAR-10 test set using different activation functions.

| Activation Function | ResNet-50 | ResNet-101 | ResNet-152 | ResNet-200 |
| --- | --- | --- | --- | --- |
| NLReLU | 82.79±0.17 ($\beta = 1.10$) | **83.47±0.28** ($\beta = 1.10$) | **83.89±0.23** ($\beta = 1.10$) | **84.12±0.31** ($\beta = 1.10$) |
| ReLU | 81.67±0.34 | 82.08±0.43 | 82.45±0.54 | 82.68±0.27 |
| Softplus | 81.46±0.31 | 81.94±0.37 | 82.23±0.28 | 82.41±0.41 |
| Swish | 82.02±0.27 | 82.50±0.13 | 82.77±0.38 | 83.09±0.34 |
| LReLU | 81.39±0.23 | 82.13±0.14 | 82.42±0.31 | 82.57±0.35 |
| PReLU | 81.63±0.48 | 82.24±0.33 | 82.51±0.64 | 82.45±0.43 |
| ELU | **82.95±0.19** | 83.39±0.09 | 83.41±0.26 | 83.66±0.29 |
| SELU | 82.62±0.22 | 83.41±0.20 | 83.67±0.33 | 83.84±0.38 |

However, we find that some BN layers do not need NLReLU to activate. Experiments show that removing NLReLU from some positions allows ResNet to produce better results when using NLReLU as the activation function. Therefore, we conducted an experiment to explore from which position NLReLU should be removed. According to whether the three kinds of activation functions classified by location exist, we arranged and combined them into 8 cases. We conducted experiments with ResNet-50 in these different cases. The experimental results are listed in Table 3.

As shown in Table 3, when NLReLUs behind all BN layers are removed, i.e., case (0,0,0), the accuracy drops significantly, compared to case (1,1,1). When only the NLReLU at one position is retained, i.e., cases (1,0,0), (0,1,0) and (0,0,1), the accuracy is higher to different degrees, and the model performs best in the case of (0,1,0). In general, the model performs best in the case of (0,1,0), i.e., when NLReLU is used in ResNet, while the NLReLU at positions A and C should be removed.

Based on the conclusions above, we conducted a comparative experiment using ResNets with different depths and different activation functions with the CIFAR-10 dataset. All the experiment results are summarized in Table 4. As shown in Table 4, NLReLU provides 1.35% higher accuracy than ReLU on average and exceeds all other well-known activation functions, except ELU and SELU. The value of $\beta$ for NLReLU in ResNet should be approximately 1.10. The use of LReLU and PReLU in ResNets do not provide stable performance. NLReLU is comparable to ELU and SELU on ResNets. Swish performs better in deep convolutional neural networks than in shallow convolutional neural networks.

## V. CONCLUSIONS

An improved activation function, i.e., NLReLU, based on a parametric logarithmic transformation is proposed in this paper. Our results show that NLReLU can reduce heteroscedasticity in the data distribution among layers and the bias shift effect, thus alleviating the "dying ReLU" and vanishing gradient problems to some extent while retaining sparse activation. This speed up the learning process and improves network convergence. We conducted experiments using convolutional neural networks with ten different depths and two different datasets. The results from these experiments show that NLReLU provides improved accuracy and convergence over ReLU, and that NLReLU is comparable to other fairly-known activation functions in convolutional neural networks with different depths.